**LTV-CTDNet: Compositional Turning Decomposition for Short-Term Turning-Movement Forecasting**

**Md Atiqur Rahman Mallick***
Graduate Student, Department of Electrical and Computer Engineering
Tennessee State University, Nashville, TN, 37209
Email: mmallick@tnstate.edu

**Kamrul Hasan, PhD**
Assistant Professor, Department of Electrical and Computer Engineering
Tennessee State University, Nashville, TN, 37209
Email: mhasan1@tnstate.edu

**Robert T. White**
TSMO Manager
Nashville Department of Transportation & Multimodal Infrastructure
Nashville, TN 37210
Email: Robert.White@nashville.gov

*Corresponding Author

## ABSTRACT

**Objectives:** Short-term turning-movement forecasts can support signal control and corridor operations, yet unconstrained neural networks may produce physically impossible negative counts or movement predictions that are not explicitly tied to an approach-demand total. This study introduces a forecasting framework designed to combine competitive accuracy with structurally admissible outputs.

**Methods**: The Linear Temporal-Variable Compositional Turning Decomposition Network (LTV-CTDNet) was evaluated using seven months of 15-minute light detection and ranging (LiDAR) observations from eight monitored corridor locations in Nashville, Tennessee. Its lightweight encoder jointly processes flattened recent turning-movement history, weekly time-slot embeddings, and location embeddings. The Compositional Turning Decomposition framework separately predicts nonnegative approach totals and within-approach turning proportions. Movement forecasts are reconstructed from these components, guaranteeing nonnegative outputs and exact agreement between each model-predicted approach total and the sum of its component movements.

**Findings**: Among the evaluated predefined configurations, LTV-CTDNet achieved the lowest movement-level mean absolute error of 1.8189 and root mean squared error of 3.8072. Although its accuracy improvements over the strongest sequence models were modest, LTV-CTDNet produced no negative forecasts. In contrast, the audited unconstrained learned models generated negative values in approximately 10.6%–29.2% of their raw forecast cells.

**Novelty**: Unlike standard models that predict turning movements directly, CTD separates approach-demand magnitude from directional turning allocation and enforces nonnegativity and exact demand-allocation closure through the model architecture.

**Practical Applications:** The framework can provide movement forecasts that are directly interpretable and structurally suitable for downstream traffic analysis, proactive monitoring, and signal-management applications without clipping or coherence correction.

**INTRODUCTION**

Accurate short-term turning-movement forecasting is important for proactive traffic management because signal timing, capacity analysis, queue management, and corridor coordination depend on how approaching vehicles are distributed among left-turn, through, and right-turn movements (Rouphail et al. 1997). These data are more difficult to collect and forecast than aggregate approach volumes because each vehicle must be associated with both an entry approach and an exit direction. A conventional four-leg intersection contains 12 directional movement streams whose magnitudes and temporal patterns may differ considerably. Reliable forecasts may therefore support predictive signal control, operational analysis, congestion management, and traveler information (Ghanim and Shaaban 2019; Zhang et al. 2024; Zhang et al. 2026).

Traffic forecasting has progressed from statistical time-series methods toward machine-learning and deep-learning approaches capable of representing nonlinear, nonstationary, and interacting traffic processes (Sayed et al. 2023). At the intersection level, recurrent, graph-based, feature-fusion, physics-guided, signal-control-refined, and pretrained architectures have been used to model temporal dynamics, spatial relationships, roadway characteristics, and external influences (Shu et al. 2022; Gong et al. 2024; Xu et al. 2023; Li et al. 2025; Pan et al. 2025; Zhang et al. 2024; Zhang et al. 2026). Recent research has also moved toward shared learning across multiple intersections, allowing related locations to contribute to a common forecasting model while location-specific information preserves local differences. This direction is particularly valuable when observations are limited at individual locations, although forecasting performance remains sensitive to missing data, sample duration, and spatial coverage (Zhang et al. 2024; Zheng et al. 2024).

Despite this progress, the reviewed turning-movement studies have primarily emphasized predictive accuracy. Their models generally produce movement volumes as direct continuous outputs from temporal or spatiotemporal representations and do not explicitly separate the total demand entering an approach from the proportions allocated among left-turn, through, and right-turn movements. As a result, these output structures do not inherently enforce nonnegative movement counts or exact agreement between an internally predicted approach demand and its component movements. Negative vehicle counts are physically impossible and require correction before forecasts can be used in traffic-analysis or signal-management applications.

This issue differs from temporal decomposition, which separates a traffic time series into additive components before forecasting and recombination (Tian et al. 2025). It is more closely related to forecast coherence, in which predictions are required to satisfy known aggregation relationships. Reconciliation methods have been developed to produce coherent forecasts across hierarchical structures, and transportation studies have applied these ideas to spatially and temporally aggregated demand predictions (Athanasopoulos et al. 2024; Khalesian et al. 2024). However, coherence and nonnegativity are separate requirements. For turning-movement forecasting, the predicted movements should be nonnegative and should sum exactly to the corresponding model-predicted approach demand.

This study addresses these requirements through Compositional Turning Decomposition (CTD). CTD separately predicts a nonnegative demand magnitude for each approach and within-approach turning proportions constrained to sum to one. The left-turn, through, and right-turn forecasts are reconstructed by multiplying the predicted approach demand by their corresponding proportions. This design guarantees nonnegative movement forecasts and exact internal demand-allocation closure by construction.

The primary implementation, termed the Linear Temporal-Variable Compositional Turning Decomposition Network (LTV-CTDNet), combines CTD with shared multi-location learning. Its lightweight encoder jointly processes flattened recent turning-movement history, weekly time-slot embeddings, and location embeddings. The framework is evaluated using seven months of 15-minute LiDAR-derived observations from eight instrumented locations along an urban corridor in Nashville, Tennessee. The benchmark includes operational, linear, recurrent, Transformer-based, spatiotemporal, and zero-shot foundation models. The evaluation examines whether LTV-CTDNet can provide competitive next-15-minute forecasting accuracy while preventing negative movement-count predictions.

Training stability across five random seeds and performance under a common hyperparameter-search grid are also examined.

The study makes four contributions:

1. A shared multi-location forecasting framework that incorporates weekly time-slot and location embeddings.
2. A compositional output architecture that separates approach-demand magnitude from directional turning allocation.
3. A broad comparison with operational, linear, recurrent, Transformer-based, spatiotemporal, and zero-shot forecasting models.
4. An evaluation that considers forecasting accuracy, raw-output nonnegativity, training stability, and tuning fairness as complementary dimensions of model performance.

These contributions position the study at the intersection of turning-movement forecasting, shared multi-location learning, time-series model design, and coherent forecasting. The following section reviews the literature that motivates the specific forecasting, Method, Results, Discussion, and Conclusion.

## LITERATURE REVIEW

### From Turning-Movement Estimation to Forecasting

The literature begins with the difficulty of obtaining movement-level data. Ghanim and Shaaban (2019) treated turning movements as an estimation problem and trained an artificial neural network to infer them from approach volumes. Their work established that aggregate demand contains learnable information about directional allocation, but it did not forecast future movements from historical sequences. Zhang et al. (2024) advanced the problem to real-time forecasting by combining a stacked bidirectional LSTM with a multilayer perceptron. Using observations from 22 intersections, their PB-LSTM framework modeled recent dynamics and longer-term contextual patterns. Its use of intersection and geometric information also indicated that turning-movement forecasting could be learned jointly across locations rather than through a separate model for every intersection. Mallick (2025) applied a bidirectional LSTM to LiDAR-derived intersection counts and explicitly accounted for the frequent zero-volume intervals that characterize movement-level data.

### From Temporal Models to Shared Spatiotemporal Learning

Later studies broadened the task from temporal sequence learning to explicit spatial and contextual modeling. Gong et al. (2024) proposed a parallel adaptive feature-fusion framework that combined historical turning flows, temporal variables, weather, geographical information, graph convolution, and recurrent learning across 67 intersections. Li et al. (2025) similarly integrated residual graph convolution, convolutional processing, adaptive fusion, and residual LSTM components. Palit and Osman (2024) used multiple graphs to represent movement relationships along an arterial corridor, while Xu et al. (2023) incorporated signal-control effects and dynamic movement relationships into an arterial-network graph model. Pan et al. (2025) incorporated intersection characteristics and traffic-flow relationships through a physics-guided spatiotemporal graph framework. Rahman et al. (2022) extended graph-convolutional recurrent learning to network-wide movement-volume prediction using traffic, demand, and built-environment information. Together, these studies show a progression from isolated time-series forecasting toward shared models that learn interactions among movements, intersections, and contextual variables.

Shared learning is particularly attractive when records at individual locations are limited. However, pooling data does not automatically improve prediction. Zheng et al. (2024) found that missingness, observation duration, and spatial coverage affect forecasting accuracy; broader coverage can provide useful information but may also introduce noise. This evidence supports shared corridor learning that includes explicit spatial and temporal representations rather than treating all observations as interchangeable.

**Expanding the Benchmarking Landscape**

The broader time-series literature shows why movement forecasting should be evaluated against diverse model families. T-GCN combines graph convolution with GRU recurrence to capture spatial and temporal dependence (Zhao et al. 2020). STGCN combines graph convolution with temporal convolution to model spatial and temporal traffic dependencies (Yu et al. 2017). DLinear demonstrates that simple direct linear models can outperform complex Transformers on some benchmarks (Zeng et al. 2023). TSMixer uses MLP-based mixing across temporal and variable dimensions (Chen et al. 2023), while iTransformer treats variables as tokens so attention captures cross-variable relationships (Liu et al. 2024). STAEformer shows that adaptive spatiotemporal embeddings can make a standard Transformer highly competitive for traffic forecasting (Liu et al. 2023), and TimeXer illustrates the value of explicitly incorporating exogenous information into time-series forecasting (Wang et al. 2024). Chronos introduced pretrained time-series models for zero-shot forecasting (Ansari et al. 2024), while Chronos-2 extended this line of work from univariate to universal forecasting (Ansari et al. 2025). Zhang et al. (2026) evaluated TimesFM under few-shot and zero-shot turning-flow conditions and augmented it with decomposition and external-variable modeling. These studies suggest that complexity or pretraining alone does not establish superiority; performance must be tested on the target corridor against operational, linear, recurrent, graph-based, Transformer-based, and foundation baselines.

**From Accuracy to Structural Admissibility**

Although the reviewed turning-movement models increasingly capture temporal patterns, spatial interactions, external variables, and transfer across locations, they generally formulate movements as direct continuous predictions and evaluate them mainly through error metrics. Their output structures do not explicitly encode an approach total and its allocation among left-turn, through, and right-turn movements. Thus, predictive improvement does not ensure nonnegative outputs or exact agreement between movement sums and approach demand (Gong et al. 2024; Li et al. 2025; Zhang et al. 2024). Temporal decomposition does not directly solve this problem. Tian et al. (2025) separated traffic series into additive components with different frequency and randomness characteristics, forecast them separately, and recombined them. That approach restructures variation through time rather than enforcing relationships among simultaneous movement forecasts.

Forecast reconciliation provides the closer conceptual foundation. Athanasopoulos et al. (2024) describe methods that adjust or construct forecasts so they satisfy hierarchical aggregation constraints, including nonnegative reconciliation. Khalesian et al. (2024) applied deep learning and hierarchical reconciliation to traffic demand and improved coherence across spatial aggregation levels. At the intersection level, Mallick et al. (2026) introduced a hierarchical flow decomposition for turning-movement prediction at signalized intersections that incorporated flow conservation into the forecasting structure. Yet turning movements require a specialized form of coherence: an approach represents the total magnitude, while turning proportions represent its directional composition. The unresolved gap is therefore not simply a need for another forecasting backbone, but for an output formulation that combines shared multivariate learning with guaranteed nonnegativity and exact demand-allocation closure. This gap directly motivates LTV-CTDNet.

## METHODS

**Study Objective and Problem Formulation**

The objective of this study is to forecast the 12 turning-movement counts at each monitored location for the next 15-minute interval. A single shared model is trained on observations from all eight locations, enabling the learning of common corridor-level temporal patterns while retaining location-specific differences via a location identifier.

Let $i \in \{1, \ldots, 8\}$ denote a monitored location and let $t$ denote a 15-minute interval. At each location and interval, the observed traffic state is represented by

$$x_{i,t} = x_{i,t,1}, x_{i,t,2}, \dots, x_{i,t,12}{}^{\top} \in \mathbb{R}^{12}_{\geq 0}. \qquad (1)$$

Here:

- $i$ identifies the location.
- $t$ identifies the 15-minute interval.
- J ∈{1,...,12} Identify the 12 movement channels.
- $x_{i,t,j}$ is the observed vehicle count for the movement channel $j$.
- $x_{i,t}$ is the complete vector containing all 12 movement counts.
- $\mathbb{R}^{12}_{\geq 0}$ means that the vector has 12 entries, each nonnegative.

The 12 movement channels follow the fixed order

$$n_{e,}, n_s, n_w, s_{e,}, s_n, s_w, e_n, e_s, e_w, w_{n,}, w_e, w_s$$

For each forecasting sample, the model uses the 12 most recent 15-minute intervals:

$$X_{i,t} = \left[x_{i,t-11}, x_{i,t-10}, \dots, x_{i,t}\right]^{\top} \in \mathbb{R}^{12\times 12}_{\geq 0}. \qquad (2)$$

The matrix $X_{i,t}$ therefore contains three hours of recent traffic history. Its 12 rows represent the 12 historical intervals, and its 12 columns represent the 12 movement channels.
The forecasting task is

$$\hat{x}_{i,t+1} = f_{\theta}\left(X_{i,t}, i, s_{t+1}\right) \qquad (3)$$

where $\hat{x}_{i,t+1}$ is the predicted 12-movement vector for the next 15-minute interval, $f_{\theta}\left(\left(X_{i,t}, i, s_{t+1}\right)\right)$is the forecasting model, and $s_{t+1}$ is the weekly time slot identifier for the target interval.

**Study Corridor, Dataset, and Chronological Partitioning**

The study used traffic-count data collected from eight roadside LiDAR monitoring locations along the Clarksville Pike and Buchanan Street corridors in Nashville, Tennessee. The data were obtained through the BlueCity platform, with access provided by the Nashville Department of Transportation. The observation period extended from September 1, 2024, through March 31, 2025. Traffic counts were recorded every 15 minutes. At each monitored location, the dataset contained 20,352 timestamps and the same 12 movement-count channels define previous section.

Before forecasting samples were created, the observations at each location were arranged in chronological order. A sliding-window procedure was then applied separately to each location. For every sample, the preceding 12 intervals were used as the model input, and the immediately following interval was used as the prediction target. Because each interval represents 15 minutes, the input window covered three hours of traffic history, and the forecast horizon was the next 15 minutes.
The sample-generation process can be written as

$$\left[x_{i,t-11}, \dots, x_{i,t}\right] \longrightarrow x_{i,t+1} \qquad (4)$$

where the 12 vectors on the left form the historical input and $x_{i,t+1}$ is the target movement-count vector. Windows were created independently within each location. Therefore, an input sequence from one location was never combined with observations from another location. After the windows had been generated, samples from all eight locations were pooled for training the shared forecasting models.

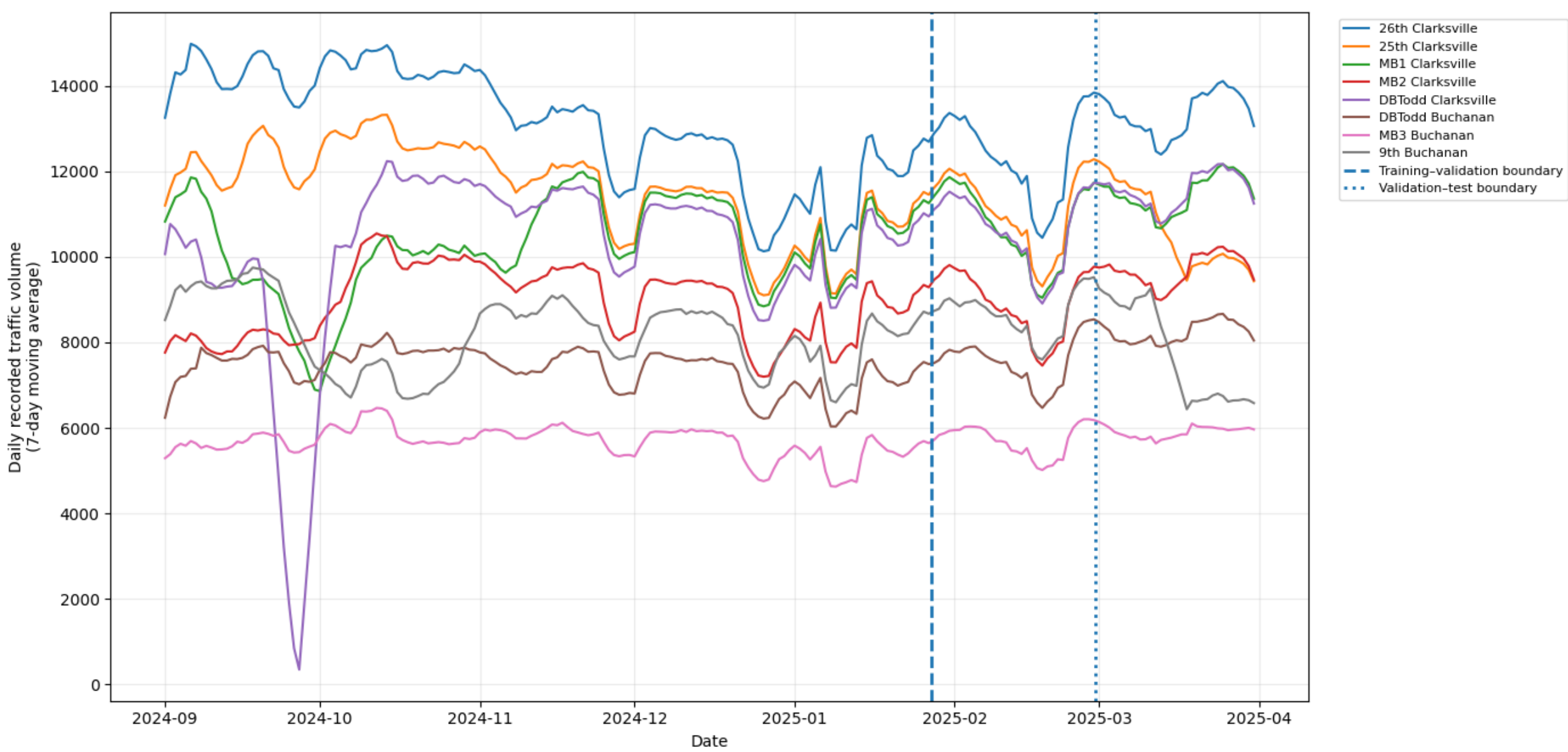

**Figure 1. Daily traffic volumes across the eight monitored corridor locations and the chronological boundaries of the training, validation, and test partitions.**

Here, Figure 1 presents the daily traffic volumes at the eight monitored locations together with the chronological partition boundaries. The first vertical marker separates the training and validation periods, while the second marker separates the validation and test periods. The training partition was used to estimate model parameters and preprocessing statistics. The validation partition was used for early stopping and hyperparameter selection. The test partition remained unseen during model development and was used only for the final evaluation.

### LTV-CTDNet Architecture and CTD Formulation

Figure 2 presents the overall architecture of LTV-CTDNet. The model consists of an LTV encoder followed by the CTD framework. The encoder combines the historical traffic lookback with weekly and Location embeddings to produce a latent representation. The CTD framework then generates approach totals and turning proportions, which are combined to obtain the final movement forecasts.

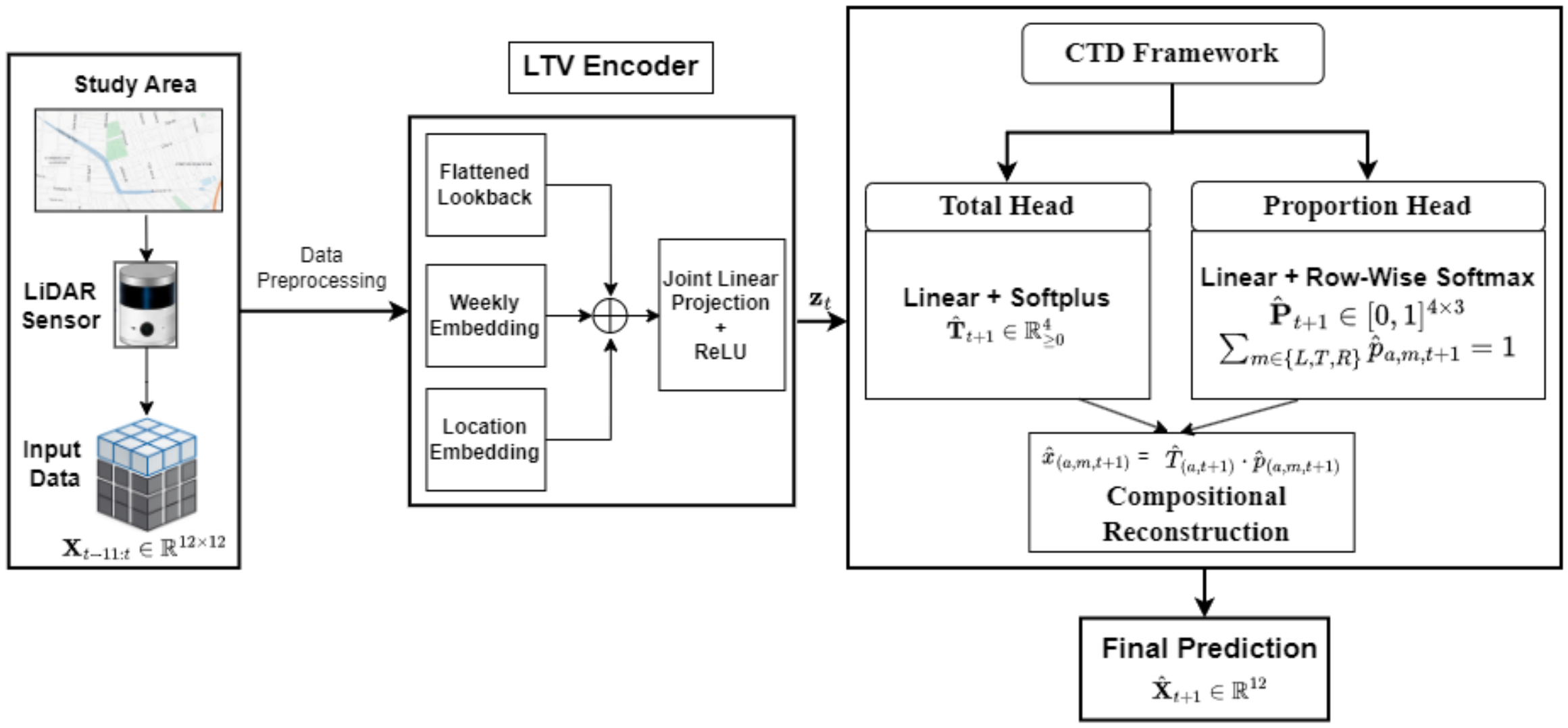

**Figure 2 Architecture of the Proposed LTV-CTDNet**

*Input Standardization and Contextual Features*

The movement channels have different traffic-volume ranges. Therefore, each input channel was standardized separately:

$$\tilde{x}_{i,t,j} = \frac{x_{i,t,j} - \mu_j}{\sigma_j}, \qquad (5)$$

where $x_{i,t,j}$ is the observed count for the movement channel $j$ at location $i$ and interval $t$, and $\mu_j$ and $\sigma_j$ are the mean and standard deviation of the channel $j$, calculated from the training inputs. The same transformation was applied to the validation and test inputs. The prediction targets were kept in raw vehicle counts.

In addition to the standardized traffic history, the model uses a weekly time-slot identifier and a location identifier. The weekly identifier represents the day of the week and a 15-minute period of the target interval, producing 672 possible slots. Each slot is represented by a learned 16-dimensional embedding. The location identifier is represented by a learned eight-dimensional embedding, allowing the shared model to retain location-specific characteristics.

**LTV Encoder**

The LTV encoder, shown in the middle block of Figure 2, combines three inputs: the flattened standardized traffic lookback, a learned weekly embedding, and a learned location embedding. These representations are concatenated and passed through a joint linear projection with ReLU activation to produce the latent representation used by the CTD framework.

The $12 \times 12$ lookback is flattened into a 144-dimensional vector and concatenated with the 16-dimensional weekly embedding and eight-dimensional location embedding:

$$u_{i,t} = \text{vec}(\tilde{X}_{i,t}) \oplus e^{w}_{s_{t+1}} \oplus e^{s}_{i} \in \mathbb{R}^{168}. \qquad (6)$$

where $\text{vec}(\tilde{X}_{i,t})$ is the flattened standardized lookback, $\oplus$ denotes concatenation, $\boldsymbol{e}^{w}_{s_{t+1}}$ is the weekly embedding (superscript $w$ means weekly), and $\boldsymbol{e}^{s}_{i}$ is the location embedding (superscript $s$ means location), $u_{i,t}$ is the combined encoder input.

The concatenated input is then passed through one fully connected layer with ReLU activation:

$$z_{i,t} = \text{ReLU}(W_e u_{i,t} + b_e) \in \mathbb{R}^{128}. \qquad (7)$$

where $z_{i,t}$ is the latent representation produced by the LTV encoder, $W_e$ the encoder's learnable weight matrix, $b_e$ the encoder's learnable bias vector. Because the historical input is flattened before projection, the encoder jointly processes information across both time intervals and movement channels.

**CTD Framework**

The latent representation produced by the LTV encoder, $z_{i,t} \in \mathbb{R}^{128}$, is passed to two parallel output heads: the Total Head and the Proportion Head. The Total Head predicts the amount of traffic entering each approach, while the Proportion Head predicts how that traffic is distributed among the three movements within each approach.

Let $a \in \{1, \dots, 4\}$ index the four approaches, and let $m \in \mathcal{M} = \{L, T, R\}$ index the left-turn, through, and right-turn movements within an approach.

**Total Head**

The Total Head predicts one total traffic value for each of the four approaches:

$$\hat{T}_{i,t+1} = \text{Softplus}(W_T z_{i,t} + b_T) \in \mathbb{R}^{4}_{\geq 0}, \qquad (8)$$

where $\hat{T}_{i,t+1}$ is the vector of predicted approach totals for location $i$ at target interval $t+1$. The matrix

$$W_T \in \mathbb{R}^{4 \times 128}$$

is the learnable weight matrix of the Total Head, and

$$b_T \in \mathbb{R}^4$$

is its learnable bias vector. The scalar $\hat{T}_{i,a,t+1}$ denotes the predicted total traffic demand for approach $a$.

The Softplus activation is applied elementwise and ensures that every predicted approach total is nonnegative.

**Proportion Head**

The Proportion Head first produces 12 unconstrained values:

$$W_P z_{i,t} + b_P,$$

where

$$W_P \in \mathbb{R}^{12\times128}$$

is the learnable weight matrix of the Proportion Head and

$$b_P \in \mathbb{R}^{12}$$

is its learnable bias vector.

These 12 values are reshaped into a $4 \times 3$ logit matrix:

$$G_{i,t+1} = \text{reshape}_{4\times3}\left(W_P z_{i,t} + b_P\right) \in \mathbb{R}^{4\times3}, \qquad (9)$$

where $G_{i,t+1}$ is the full logit matrix. Each row corresponds to one approach, and the three columns correspond to the left-turn, through, and right-turn movements. The scalar $G_{i,a,m,t+1}$ denotes the logit for movement $m$ of approach $a$.

A separate Softmax is applied to each row:

$$\hat{p}_{i,a,m,t+1} = \frac{\exp\left(G_{i,a,m,t+1}\right)}{\sum_{r\in\mathcal{M}} \exp\left(G_{i,a,r,t+1}\right)}, \qquad (10)$$

where $r \in \mathcal{M}$ is a summation index over the three movements $L$, $T$, and $R$. The scalar $\hat{p}_{i,a,m,t+1}$ is the predicted proportion assigned to movement $m$ of approach $a$.

The complete proportion matrix is

$$\hat{P}_{i,t+1} = \left[\hat{p}_{i,a,m,t+1}\right] \in [0,1]^{4\times3}. \qquad (11)$$

Because Softmax is applied separately to each approach row, the three predicted proportions within each approach sum to one:

$$\sum_{m\in\mathcal{M}} \hat{p}_{i,a,m,t+1} = 1,\ a \in \{1,\dots,4\}. \qquad (12)$$

*Compositional Reconstruction*

The predicted traffic count for movement $m$of approach $a$ is obtained by multiplying the predicted approach total by the corresponding predicted proportion:

$$\hat{x}_{i,a,m,t+1} = \hat{T}_{i,a,t+1} \cdot \hat{p}_{i,a,m,t+1} \qquad (13)$$

where $\hat{x}_{i,a,m,t+1}$is the predicted vehicle count for movement $m$ of approach $a$ at location $i$during interval $t+1$.

The 12 reconstructed movement counts are then arranged into the final prediction vector:

$$\hat{x}_{i,t+1} \in \mathbb{R}^{12}_{\geq 0}. \qquad (14)$$

This formulation provides two structural guarantees. First, every predicted movement count is nonnegative:

$$\hat{x}_{i,a,m,t+1} \geq 0. \qquad (15)$$

This follows because both $\hat{T}_{i,a,t+1}$and $\hat{p}_{i,a,m,t+1}$are nonnegative.

Second, the three predicted movement counts associated with each approach sum exactly to the predicted total for that approach:

$$\begin{aligned}\sum_{m\in\mathcal{M}} \hat{x}_{i,a,m,t+1} &= \sum_{m\in\mathcal{M}} \hat{T}_{i,a,t+1}\, \hat{p}_{i,a,m,t+1} \\ &= \hat{T}_{i,a,t+1} \sum_{m\in\mathcal{M}} \hat{p}_{i,a,m,t+1} \qquad (16) \\ &= \hat{T}_{i,a,t+1}.\end{aligned}$$

Therefore, nonnegativity and exact approach-level closure are enforced directly by the CTD architecture. No clipping, post-processing, or additional constraint penalty is required.

**Benchmark Models**

The proposed method is compared with approaches spanning operational, linear, recurrent, attention-based, spatial, and foundation-model categories.

Persistence predicts that the next state equals the most recent observation. DirectLinear maps the flattened three-hour history directly to the twelve outputs. The sequence-learning baselines are a gated recurrent unit (Shu et al. 2022), a long short-term memory network (Zhang et al. 2024), a conventional Transformer encoder included as a standard time-series benchmark by Zeng et al. (2023), and an iTransformer (Liu et al. 2024) with reversible instance normalization (RevIN) applied per window on raw inputs, a component on which its published performance depends. The spatial baselines are STGCN (Yu et al. 2017) and STAEformer (Liu et al. 2023), trained on the panel of all eight locations. STGCN represents the locations as nodes of a chain graph in corridor order (adjacent sensors connected, with self-loops), while STAEformer applies temporal and spatial attention to the panel. Two zero-shot foundation models complete the suite: Chronos-Bolt (Ansari et al. 2024), forecasting each movement univariately, and Chronos-2 (Ansari et al. 2025), processing the twelve movements jointly. Both use 96 historical intervals (24 hours) of context, take the median quantile as the point forecast, and require no study-specific training; they represent the pretrained zero-shot forecasting category recently examined for turning-flow prediction (Zhang et al. 2026).

Output biases of the unconstrained supervised baselines were initialized from the training-set movement means; the proposed model correspondingly initialized its approach-total head from the training-set mean approach totals, so every trained model begins from the same climatological forecast.

**Training and Hyperparameter-Selection Protocol**

All supervised models were trained using movement-level mean squared error:

$$\mathcal{L}_{\text{MSE}} = \frac{1}{BM} \sum_{b=1}^{B} \sum_{j=1}^{M} \left(x_{b,j} - \hat{x}_{b,j}\right)^2, \qquad (17)$$

where $B = 64$ is the batch size and $M = 12$ is the number of movement channels, $b$ and $j$ are just index variables inside the loss function. No auxiliary loss was applied to the CTD outputs because nonnegativity and approach-level closure are enforced directly by the architecture.

Training used the Adam optimizer with an initial learning rate of $10^{-3}$, weight decay of $10^{-3}$, and a maximum of 100 epochs. Early stopping was applied after 10 consecutive epochs without improvement in validation MSE, and the best validation weights were restored for evaluation. Each trainable model was trained with five random seeds $\{0,1,2,3,4\}$, and results are reported as mean.

Hyperparameters were selected using the training and validation partitions only. The unified search crossed representation widths $\{64,128\}$, depths $\{1,2\}$, and learning rates $\{10^{-3}, 3 \times 10^{-4}\}$, resulting in eight configurations per configurable model. Architecture-specific eight-configuration grids were used for STGCN and STAEformer.

Each candidate configuration was trained on the training partition and ranked by validation MAE. The test set was not used for model selection. The selected configuration for each model was then trained under the five-seed protocol and evaluated on the held-out test set. The primary results use predefined model configurations, while a separate tuned comparison reports the validation-selected configurations. The two comparisons are reported separately.

The benchmark models were implemented using their standard formulations, while LTV-CTDNet was evaluated as a complete framework. Accordingly, the forecasting comparison represents end-to-end performance across the evaluated model formulations. The nonnegativity and exact demand-allocation closure properties are provided specifically by the CTD output structure.

**Evaluation Metrics and Admissibility Audit**

Forecasting performance was evaluated on the held-out test set using movement-level mean absolute error (MAE),  and root mean squared error (RMSE).

$$\text{MAE} = \frac{1}{NM} \sum_{n=1}^{N} \sum_{j=1}^{M} |x_{n,j} - \hat{x}_{n,j}|, \qquad (18)$$

$$\text{RMSE} = \sqrt{\frac{1}{NM} \sum_{n=1}^{N} \sum_{j=1}^{M} \left(x_{n,j} - \hat{x}_{n,j}\right)^2}. \qquad (19)$$

Here, $n$ indexes a test sample, $N$is the total number of test samples, $j$ indexes a movement channel, and $M = 12$. The terms $x_{n,j}$and $\hat{x}_{n,j}$denote the observed and predicted counts, respectively. MAE measures the average absolute prediction error, while RMSE gives greater weight to larger errors. MAPE was not used because the dataset contains many zero and low-volume observations.

Physical admissibility was evaluated by reporting the percentage of negative raw predictions and the most negative predicted value. These checks were applied before clipping or other corrective post-processing.

**RESULTS**

Here, (**Table 1**) presents the next-15-minute forecasting results for the predefined model configurations. Metrics for supervised models are reported as the mean across five independently trained seeds. Persistence and the zero-shot foundation models are deterministic under the implemented evaluation and are therefore reported as single values.

LTV-CTDNet achieved the lowest MAE, and RMSE among the evaluated models. Relative to persistence, it reduced MAE by 17.5%, and RMSE by 18.4%. This indicates that the model captured short-term demand changes beyond simply repeating the most recent observation.

The linear baselines improved on persistence, confirming that the recent three-hour history contains useful predictive information. However, their errors remained higher than those of the recurrent, attention-based, and proposed models.

**TABLE 1 Corridor-Wide Next-15-Minute Forecasting Performance**

| Model | MAE | RMSE | Parameters |
|---|---|---|---|
| Persistence | 2.2038 | 4.6632 | 0 |
| Direct Linear | 1.9764 | 4.1775 | 1,740 |
| GRU | 1.8811 | 3.8518 | 15,756 |
| LSTM | 1.8691 | 3.8361 | 20,748 |
| Transformer | 1.8624 | 3.8141 | 69,324 |
| iTransformer (RevIN) | 1.9204 | 4.1001 | 67,841 |
| STGCN | 2.6023 | 5.9726 | 27,628 |
| STAEformer | 1.9030 | 4.0686 | 89,348 |
| Chronos-Bolt, zero-shot | 1.9307 | 4.2539 | — |
| Chronos-2, zero-shot | 1.8675 | 4.2019 | — |
| **LTV-CTDNet** | **1.8189** | **3.8072** | **34,512** |

Among the conventional sequence models, the Transformer produced the strongest baseline MAE of 1.8624. LTV-CTDNet reduced that MAE by 2.3%. The RMSE values were close to 3.8072 for LTV-CTDNet and 3.8141 for the Transformer, suggesting that their large-error behavior was similar, although LTV-CTDNet retained a small advantage.

Persistence produced an MAE of 2.2038 and an RMSE of 4.6632, establishing the operational benchmark for repeating the most recent observation. Direct Linear improved these values to an MAE of 1.9764 and an RMSE of 4.1775, indicating that the flattened three-hour history contained useful information beyond the latest observation alone. However, both models were less accurate than the recurrent, attention-based, and proposed approaches.

The recurrent models were also competitive. LSTM achieved an MAE of 1.8691, followed by GRU at 1.8811. These results show that relatively standard sequence models remained strong benchmarks for the 15-minute forecasting task.

The two spatial models produced different outcomes. STAEformer achieved an MAE of 1.9030, while STGCN performed substantially worse, with an MAE of 2.6023. For this dataset and forecast horizon, the evaluated graph and panel formulations did not consistently improve prediction over simpler sequence-based models.

The zero-shot foundation models were competitive despite receiving no study-specific training. Chronos-Bolt and Chronos-2 were evaluated using pretrained checkpoints without fine-tuning; therefore, no model parameters were estimated using the study dataset. Chronos-2 achieved an MAE of 1.8675, close to the Transformer and LSTM results. However, its RMSE was higher than those of LTV-CTDNet. Relative to Chronos-2, LTV-CTDNet reduced MAE by 2.6%, and RMSE by 9.4%.The larger RMSE difference suggests that Chronos-2 experienced more high-error cases even though its average absolute error was competitive.

LTV-CTDNet used 34,512 trainable parameters. This was approximately 50% fewer than the predefined Transformer and 61% fewer than STAEformer. The model was not the smallest in the benchmark, but it provided a favorable balance among predictive accuracy, model size, and structurally admissible outputs.

**Training Stability**

LTV-CTDNet produced similar test performance across the five random seeds. The individual MAE values were 1.8185, 1.8424, 1.8188, 1.8165, and 1.7983, resulting in a standard deviation of 0.0157. The relatively small spread indicates that the reported performance was not driven by a single favorable initialization. The representative training history also showed a rapid decrease in training and validation loss during the first few epochs, followed by gradual stabilization. Here, (**Figure 3**) illustrates the training and validation MSE for a representative LTV-CTDNet run. Both curves decreased rapidly during the first few epochs and then stabilized. The lowest validation MSE occurred at epoch 16, and the corresponding model weights were restored for evaluation.

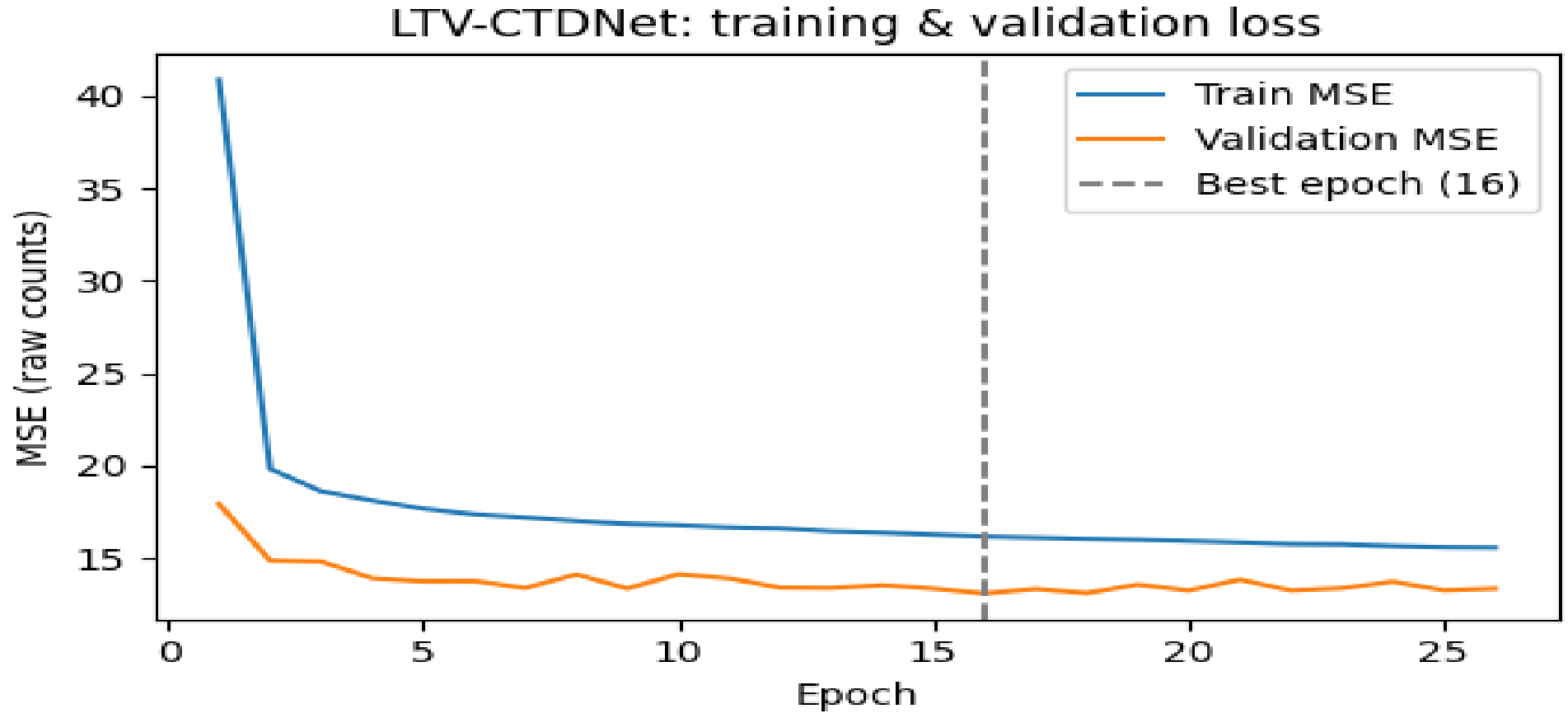

**Figure 3 LTV-CTDNet Training and Validation Loss Curve**

**Negative Forecast Check**

In addition to forecasting accuracy, this study examined whether the models produced traffic-count forecasts that could be used without corrective post-processing. This evaluation is important because unconstrained regression models may produce negative vehicle counts, which have no physical or operational meaning and must be corrected before use in traffic-analysis applications. The analysis therefore addressed the following question: Can LTV-CTDNet provide nonnegative turning-movement forecasts directly, without clipping negative outputs to zero?

Chronos-Bolt and Chronos-2 were excluded because their implemented inference pipelines clipped negative predictions before the evaluated outputs were retained. Their available predictions therefore could not be compared with the unmodified raw forecasts of the other models.

.

**TABLE 2 Admissibility Audit of Raw Forecast Outputs**

| Model | Negative Output Cells (%) | Most Negative Forecast |
|---|---|---|
| Persistence | 0.000 | 0.000 |
| DirectLinear | 12.482 | − 6.023 |
| GRU | 13.914 | −3.983 |
| LSTM | 10.638 | −3.856 |
| Transformer | 10.977 | −4.118 |
| iTransformer (RevIN) | 29.223 | −5.680 |
| STGCN | 17.056 | −22.636 |
| STAEformer | 16.086 | −1.513 |
| **LTV-CTDNet** | **0.000** | **0.000** |

The results answer the evaluation question affirmatively. Among the audited learned models, LTV-CTDNet produced no negative turning-movement forecasts, with 0.000% negative output cells and a minimum forecast of 0.000 vehicles. In contrast, the unconstrained learned models produced negative values in approximately 10.6% to 29.2% of their raw forecast cells. The iTransformer produced the highest percentage of negative predictions, whereas STGCN produced the most severe negative forecast of −22.636 vehicles.

Persistence also produced no negative values because it simply repeated the latest observed nonnegative traffic counts. Whereas LTV-CTDNet achieved nonnegative outputs for a different reason:

its CTD structure predicts nonnegative approach totals and combines them with nonnegative turning proportions. Therefore, its reconstructed movement forecasts are nonnegative by design rather than through clipping or post-processing.

**Hyperparameter Analysis**

The primary forecasting comparison in (**Table 1**) used predefined model configurations evaluated on the held-out test set. Because comparative performance can be influenced by differences in architecture size, learning rate, and tuning effort, a secondary unified-grid analysis was conducted as a model-selection sensitivity check. Each configurable neural model received the same search budget of eight configurations, formed from two representation-width settings, two architecture-specific structural settings, and two learning rates. Each configuration was trained using the training partition and seed 0. Early stopping and checkpoint restoration were based on validation mean squared error, whereas the configurations were ranked using raw-count validation mean absolute error (MAE). Here, (**Table 3**) reports on the configuration achieving the lowest validation MAE for each model. This analysis evaluates sensitivity to tuning effort and does not replace the primary held-out test comparison in (**Table 1**). Persistence, DirectLinear, Chronos-Bolt, and Chronos-2 were not included because they either did not have comparable width-and-depth hyperparameters or required no study-specific model tuning.

**TABLE 3 Best Validation Configurations Under the Unified Grid**

| Model | Selected Configuration | Validation MAE |
|---|---|---|
| GRU | hidden 128, 1 layer, learning rate 0.0003 | 1.7294 |
| LSTM | hidden 128, 1 layer, learning rate 0.0003 | 1.7351 |
| Transformer | dimension 128, 2 layers, learning rate 0.0003 | 1.7493 |
| iTransformer (RevIN) | dimension 128, 2 layers, learning rate 0.001 | 1.8645 |
| **LTV-CTDNet** | **hidden 128, 2 layers, learning rate 0.001** | **1.7226** |
| STAEformer | feature dimension 32, 2 layers, learning rate 0.0003 | 1.7420 |
| STGCN | 64 channels, kernel size 2, learning rate 0.001 | 2.0925 |

LTV-CTDNet achieved the lowest validation MAE in the unified search, with a value of 1.7226. However, GRU, LSTM, STAEformer, and Transformer also achieved similar validation performance, with MAE values between 1.7294 and 1.7493. These results indicate that LTV-CTDNet's performance was not simply caused by receiving more favorable hyperparameter tuning. Because the differences among the strongest models were small, (**Table 3**) should be interpreted as a check on tuning fairness rather than as evidence that LTV-CTDNet was substantially better than every baseline.

The two-layer LTV-CTDNet achieved a validation MAE of 1.7226, while the predefined one-layer model achieved 1.7227. This difference was negligible. Therefore, adding a second layer did not provide a meaningful improvement. The simpler one-layer model was retained for the primary comparison because it achieved nearly identical validation accuracy with lower complexity.

**DISCUSSION**

While LTV-CTDNet achieved the strongest overall forecasting performance among the evaluated predefined configurations, its improvement over strong sequence models such as Transformer and LSTM was modest. Rather than demonstrating overwhelming predictive superiority, the model's main contribution lies in combining competitive forecasting accuracy with a structurally constrained output design. By separating approach demand from turning allocation within the CTD framework, the model produces nonnegative turning-movement forecasts and preserves consistency between each predicted approach total and its component movements. Because negative vehicle counts are physically impossible, incorporating these requirements directly into the architecture avoids the need for clipping or other corrective post-processing and makes the forecasts more suitable for downstream traffic-analysis applications.

The findings also provide insight into the relationship between model complexity and short-term turning-movement forecasting performance. The competitive results of GRU, LSTM, and Transformer indicate that recent turning-movement history contains substantial information for next-15-minute prediction. In contrast, more complex spatial models such as STGCN did not provide consistent improvements under the evaluated corridor representation. This outcome may reflect the limited size of the eight-location network, the selected graph structure, or spatial relationships that were not fully represented by the implemented graph. The five-seed evaluation supports the stability of LTV-CTDNet across random initializations, while the unified-grid analysis indicates that its primary result was not solely caused by unequal tuning effort. However, the validation differences among the strongest models remained small. Because the components of LTV-CTDNet were evaluated jointly, the present study does not isolate the individual contributions of the contextual embeddings, shared encoder, and CTD structure to forecasting accuracy. Overall, the findings suggest that realistic output structure can be as important as increasing forecasting-backbone complexity. Future work should evaluate the framework across larger transportation networks, multiple forecast horizons, alternative forecasting backbones and graph structures, and fully input-matched benchmark comparisons.

## CONCLUSIONS

This study developed LTV-CTDNet for next-15-minute turning-movement forecasting using seven months of LiDAR-derived observations from eight corridor locations. The framework combines shared corridor learning with temporal and location embeddings and a compositional traffic decomposition output structure that represents each approach using total demand and turning proportions. Among the evaluated predefined configurations, LTV-CTDNet achieved the lowest movement-level MAE of 1.8189 and RMSE of 3.8072. Its improvement over strong sequence models such as Transformer and LSTM was modest, but it produced no negative vehicle-count forecasts and maintained consistency between predicted approach demand and its component turning movements without clipping or corrective post-processing.

These findings indicate that realistic output constraints can be incorporated into a lightweight forecasting architecture without sacrificing predictive performance. The five-seed evaluation supported the stability of the model, while the unified-grid analysis indicated that its result was not solely attributable to more favorable hyperparameter tuning. Nevertheless, the conclusions are limited to one corridor, eight monitored locations, seven months of data, and a single 15-minute forecast horizon. Future research should evaluate LTV-CTDNet on larger and more diverse transportation networks, test multiple forecast horizons, examine alternative forecasting backbones and graph structures, and conduct fully input-matched comparisons across models.

## ACKNOWLEDGMENTS

This study was supported by the U.S. Department of Transportation and the Nashville Department of Transportation. ChatGPT, an artificial intelligence language model developed by OpenAI, was used to support idea development, sentence revision, and review of manuscript requirements. Grammarly was used for spelling and grammar checks. All AI-assisted content was reviewed and verified by the authors, who remain fully responsible for the accuracy, interpretation, and final content of the paper.

## AUTHOR CONTRIBUTIONS

The writers affirm their contributions to the manuscript as detailed below: Study conception, analysis, interpretation of results and design: Md Atiqur Rahman Mallick, idea creation and draft manuscript preparation: Dr. Kamrul Hasan, for data collecting: Robert T. White. All authors reviewed the results and approved the final version of the manuscript.

## DECLARATION OF CONFLICTING INTERESTS

The authors declared no potential conflicts of interest with respect to the research, authorship, and/or publication of this article.